\documentclass[10pt,twocolumn,letterpaper]{article}

\usepackage{cvpr}              % To produce the CAMERA-READY version
\usepackage{graphicx}
\usepackage{amsmath}
\usepackage{amssymb}
\usepackage{booktabs}
\usepackage[table]{xcolor}
\usepackage[ruled]{algorithm2e}
\usepackage{multirow}

\usepackage{docmute}
\usepackage{chapterbib}

\newcommand{\tabincell}[2]{
\begin{tabular}{@{}#1@{}}#2\end{tabular}
}

\graphicspath{{figs/}}
\DeclareGraphicsExtensions{.pdf,.jpg,.jpeg,.png}

\usepackage[pagebackref,breaklinks,colorlinks]{hyperref}

\usepackage[capitalize]{cleveref}
\crefname{section}{Sec.}{Secs.}
\Crefname{section}{Section}{Sections}
\Crefname{table}{Table}{Tables}
\crefname{table}{Tab.}{Tabs.}

\def\cvprPaperID{605} % *** Enter the CVPR Paper ID here
\def\confName{CVPR}
\def\confYear{2022}

\begin{document}

\title{Graph Sampling Based Deep Metric Learning for\\Generalizable Person Re-Identification}

\author{Shengcai Liao\thanks{Shengcai Liao is the corresponding author.}\\
Inception Institute of Artificial Intelligence (IIAI)\\
Masdar City, Abu Dhabi, UAE\\
{\tt\small scliao@ieee.org}
\and
Ling Shao\\
Terminus Group\\
China\\
{\tt\small ling.shao@ieee.org}
}

\maketitle

%%%%%%%%% ABSTRACT
\begin{abstract}
  %Generalizable person re-identification has recently gained increasing attention due to its research as well as practical value.
  Recent studies show that, both explicit deep feature matching as well as large-scale and diverse training data can significantly improve the generalization of person re-identification. However, the efficiency of learning deep matchers on large-scale data has not yet been adequately studied. Though learning with classification parameters or class memory is a popular way, it incurs large memory and computational costs. In contrast, pairwise deep metric learning within mini batches would be a better choice. However, the most popular random sampling method, the well-known PK sampler, is not informative and efficient for deep metric learning. Though online hard example mining has improved the learning efficiency to some extent, the mining in mini batches after random sampling is still limited. This inspires us to explore the use of hard example mining earlier, in the data sampling stage. To do so, in this paper, we propose an efficient mini-batch sampling method, called graph sampling (GS), for large-scale deep metric learning. The basic idea is to build a nearest neighbor relationship graph for all classes at the beginning of each epoch. Then, each mini batch is composed of a randomly selected class and its nearest neighboring classes so as to provide informative and challenging examples for learning. Together with an adapted competitive baseline, we improve the state of the art in generalizable person re-identification significantly, by 25.1\% in Rank-1 on MSMT17 when trained on RandPerson. Besides, the proposed method also outperforms the competitive baseline, by 6.8\% in Rank-1 on CUHK03-NP when trained on MSMT17. Meanwhile, the training time is significantly reduced, from 25.4 hours to 2 hours when trained on RandPerson with 8,000 identities. Code is available at \url{https://github.com/ShengcaiLiao/QAConv}.
\end{abstract}

\begin{figure*}
  \centering
  \includegraphics[width=60mm]{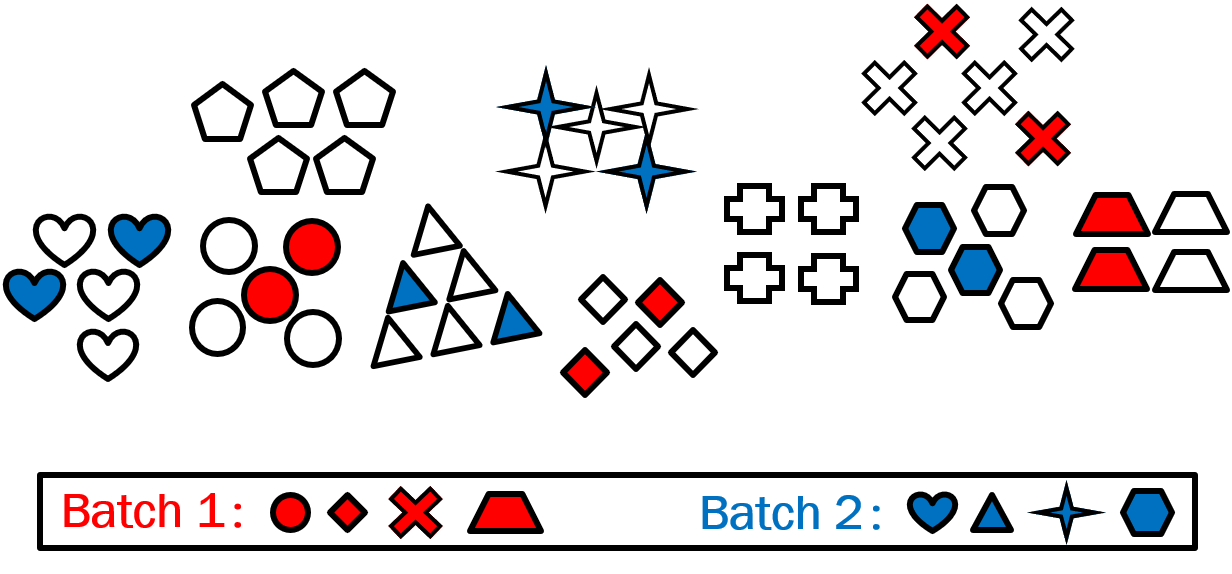}
  \hspace{10mm}
  \includegraphics[width=60mm]{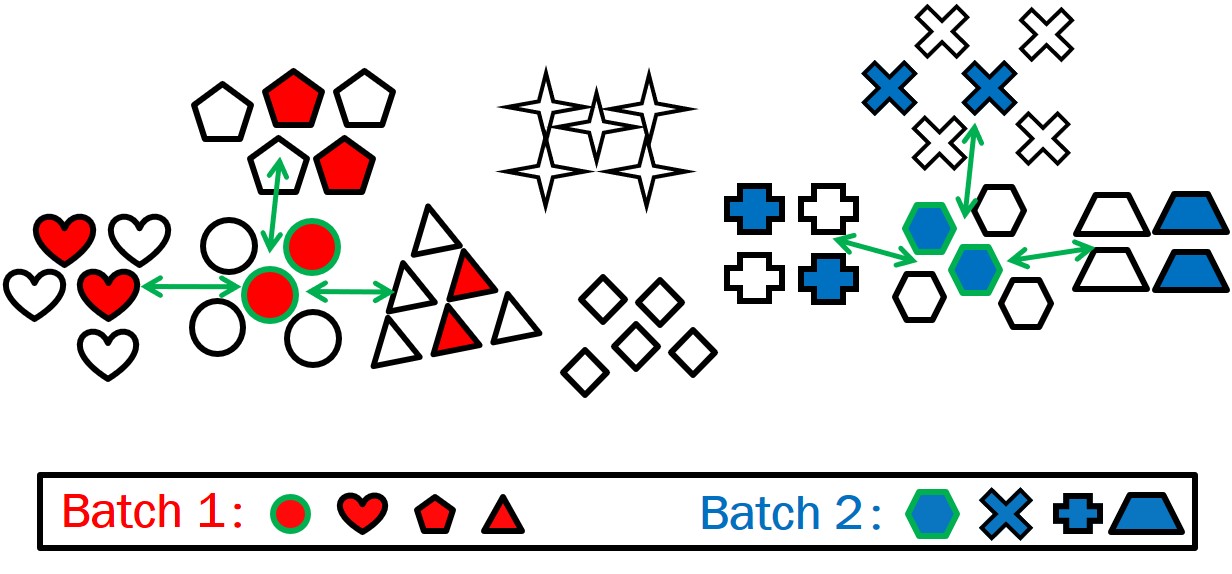}\\
  (a) PK sampler
  \hspace{50mm}
  (b) GS sampler\\
  \caption{Two different sampling methods: (a) PK sampler; and (b) the proposed GS sampler. Different shapes indicate different classes, while different colors indicate different batches. GS constructs a graph for all classes and always samples nearest neighboring classes.}\label{fig:GS}
  \end{figure*}

  %%%%%%%%% BODY TEXT
  \section{Introduction}

  Person re-identification is a popular computer vision task, where the goal is to find a person, given in a query image, from the search over a large set of gallery images. In the last two years, generalizable person re-identification has gain increasing attention due to both its research and practical value \cite{song2019generalizable,jia2019frustratingly,Zhou2019-OSNet,Qian2020-MuDeep,Jin2020-SNR,Liao-ECCV2020-QAConv,Zhuang2020-CBN}. This task studies the generalizability of a learned person re-identification model in unseen scenarios, and employs direct cross-dataset evaluation \cite{yi2014deep,Hu2014Cross} for performance benchmarking.

  For deep metric learning, beyond feature representation learning and loss designs, explicit deep feature matching schemes are shown to be effective for matching person images~\cite{ahmed2015improved,Li-CVPR-2014-DeepReID,shen2018end,suh2018part-aligned,Liao-ECCV2020-QAConv}, due to the advantages in addressing pose and viewpoint changes, occlusions, and misalignments. In particular, a recent method, called query-adaptive convolution (QAConv) \cite{Liao-ECCV2020-QAConv}, has proved that explicit convolutional matching between gallery and query feature maps is quite effective for generalizable person re-identification. However, these methods all require more computational costs compared to conventional feature learning methods.

  Beyond novel generalizable algorithms, another way to improve generalization is to enlarge the scale and diversity of the training data. For example, a recent dataset called RandPerson \cite{Wang2020-RandPerson} synthesized 8,000 identities, while \cite{wang2020-sysu30k} and \cite{bai2021person30k} both collected 30K persons for re-identification training. These studies all observed improved generalization ability for person re-identification. However, the efficiency of deep metric learning from large-scale data has not yet been adequately studied in person re-identification.

  There are some popular ways of learning deep person re-identification models, including classification (with the ID loss \cite{Zheng-CVPR2017-IDE}), metric learning (with a pairwise loss \cite{yi2014deep,Deng2018} or triplet loss \cite{Hermans2017-Triplet}), and their combinations (e.g. ID + triplet loss). Using an ID loss is convenient for classification learning. However, in large-scale deep learning, involving classifier parameters incurs large memory and computational costs in both the forward and backward passes. Similarly, involving class signatures for metric learning in a global view is also not efficient. For example, QAConv in \cite{Liao-ECCV2020-QAConv} is difficult to scale up for large-scale training, because a class memory module is designed, where full feature maps are stored for all classes as signatures, and they are required for cross feature map convolutional matching during training.

  Therefore, involving class parameters or signatures in either classification or metric learning is not efficient for large-scale person re-identification training. In contrast, we consider that pairwise deep metric learning between samples in mini batches is better suited for this task. Accordingly, the batch sampler plays an important role for efficient learning \cite{Hermans2017-Triplet,Ye2020Survey}. The well-known PK sampler \cite{schroff2015facenet,Hermans2017-Triplet} is the most popular random sampling method in person re-identification. It first randomly selects $P$ classes, and then randomly samples $K$ images per class to construct a mini batch of size $B=P\times K$. Since this is performed randomly, the sampled instances within a mini batch are uniformly distributed across the whole dataset (see Fig. \ref{fig:GS} (a)), and might therefore not be informative and efficient for deep metric learning. To address this, an online hard example mining method was proposed in \cite{Hermans2017-Triplet}, which improved the learning efficiency to some extent. However, the mining is performed online on the already sampled mini batches. Therefore, this method is still limited by the fully random PK sampler, because the mini batches obtained by this sampler do not consider the sample relationship information.

  To address this, we propose to shift the hard example mining earlier to the data sampling stage. Accordingly, we propose an efficient mini-batch sampling method, called graph sampling (GS), for large-scale deep metric learning. The basic idea is to build a nearest neighbor relationship graph for all classes at the beginning of each epoch. Then, the mini-batch sampling is performed by randomly selecting a class as anchor, and its top-k nearest neighboring classes, with the same $K$ instances per class, as shown in Fig. \ref{fig:GS} (b). This way, instances within a sampled mini batch are mostly similar to each other, so as to provide informative and challenging examples for discriminant learning. From face recognition loss function studies \cite{wen2016discriminative,liu2017sphereface,Deng-CVPR19-ArcFace}, it is known that focusing on boundary (hard) examples helps improving the discriminant ability of the learned model, and helps resulting in compact representations that generalize well beyond the training data. The GS sampler shares a similar idea in focusing on nearest neighboring classes, and thus has a potential of improving the discrimination and generalization ability of the learned model.

  In summary, the contributions of this paper include: (1) We propose a new mini-batch sampling method, termed GS, and prove that it enables more efficient learning than the well-known PK sampler; (2) We improve a very competitive baseline by 6.8\% in Rank-1 with MSMT17 $\rightarrow$ CUHK03-NP, and reduce the training time significantly, from 25.4 hours to 2 hours on RandPerson with 8,000 identities; and (3) Together with the baseline, we improve the state of the art in generalizable person re-identification significantly, by 20.6\% in Rank-1 with Market-1501 $\rightarrow$ MSMT17, and by 25.1\% in Rank-1 with RandPerson $\rightarrow$ MSMT17.

  \section{Related Work}

  Metric learning approaches have been widely studied in the early stage of person re-identification. Many algorithms have been proposed, such as the well-known PRDC \cite{DBLP:conf/cvpr/ZhengGX11}, KISSME \cite{kostinger2012large}, and XQDA \cite{Liao-CVPR-2015-LOMO-XQDA}, to name a few. In recent years, deep metric learning in particular has become popular and been extensively studied. Beyond feature representation learning, specific deep metric learning can be roughly classified in terms of loss function designs and deep feature matching schemes. For loss function designs, pairwise loss functions \cite{yi2014deep,Deng2018}, classification or identification loss \cite{Zheng-CVPR2017-IDE}, and triplet loss \cite{DBLP:conf/cvpr/ZhengGX11,schroff2015facenet,Hermans2017-Triplet} are the most popular. For deep feature matching schemes, a number of methods have been proposed in the literature. For example, Ahmed et al. proposed a deep convolutional architecture with layers specifically designed for local neighborhood matching~\cite{ahmed2015improved}. Li et al. proposed a novel filter pairing neural network (FPNN) to jointly handle several known challenges, such as misalignment and occlusions \cite{Li-CVPR-2014-DeepReID}. Shen et al. proposed an end-to-end deep Kronecker-Product Matching (KPM) network \cite{shen2018end} for softly aligned matching. Suh et al. proposed a deep neural network to learn part-aligned bilinear representations \cite{suh2018part-aligned}. Liao and Shao proposed the query-adaptive convolution (QAConv) for explicit deep feature matching, and proved its effectiveness for generalizable person re-identification \cite{Liao-ECCV2020-QAConv}. They also proposed a Transformer based method, TransMatcher \cite{Liao-NeurIPS2021-TransMatcher}, for improved performance.

  Generalizable person re-identification was initially studied in \cite{yi2014deep,Hu2014Cross}, where direct cross-dataset evaluation was proposed to benchmark algorithms. With advancements in deep learning, this task has gained increasing attention in recent years. For example, Song et al. \cite{song2019generalizable} proposed a domain-invariant mapping network with a meta-learning pipeline. Jia et al. \cite{jia2019frustratingly} adopted both instance and feature normalization to alleviate both style and content variances across datasets. Zhou et al. proposed a new backbone network called OSNet \cite{Zhou2019-OSNet}, and further demonstrated its advantages in generalizing deep models \cite{Zhou2019-OSNet}. %Qian et al. proposed a deep leader-based multi-scale attention architecture (MuDeep) for person re-identification, with improved cross-dataset performance \cite{Qian2020-MuDeep}. 
  Jin et al. proposed a style normalization and restitution module, which shows good generalizability \cite{Jin2020-SNR}. Yuan et al. proposed an adversarial domain-invariant feature learning network (ADIN), which explicitly learns to separate identity-related features from challenging variations \cite{Yuan2020-ADIN}. Zhuang et al. proposed a camera-based batch normalization (CBN) method for domain-invariant representation learning \cite{Zhuang2020-CBN}. Recently, meta-learning has also been shown to be effective for learning generalizable models. For example, Zhao et al. proposed memory-based multi-source meta-learning (M$^3$L) for generalizing to unseen domains \cite{zhao2021learning}. Choi et al. proposed the MetaBIN algorithm for meta-training the batch-instance normalization nettwork \cite{choi2021meta-bin}. Bai et al. proposed a dual-meta generalization network and a large-scale dataset called Person30K for person re-identification \cite{bai2021person30k}. In addition to the above, Wang et al. proposed a large-scale synthetic person re-identification dataset, called RandPerson, and proved that models learned from synthesized data generalize well to real-world datasets \cite{Wang2020-RandPerson}. %Following this, Zhang et al. proposed the UnrealPerson dataset for annotation-free person re-identification \cite{zhang2021unrealperson}.

  However, the generalization of current methods is still far from satisfactory for practical person re-identification. Taking face recognition as a good example in practice, future directions may gradually be learning from more large-scale data for better performance. %, in either supervised, semi-supervised, or unsupervised way. 
  However, the efficiency of large-scale learning has been inadequately studied in person re-identification. As basic as the mini-batch sampler, though it plays an important role in deep metric learning \cite{wu2017sampling-matters,Hermans2017-Triplet,Ye2020Survey}, it still has not yet been much studied.

  Beyond online hard example mining within mini batches \cite{Hermans2017-Triplet}, several methods have been proposed for hard example mining during data sampling for mini batches. Suh et al. \cite{suh2019stochastic} proposed a stochastic class-based hard example mining for deep metric learning. It uses learnable class signatures to find nearest classes, and further performs an instance-level refined search within the subset of classes found in the first stage for hard example mining. Besides, the Doppelganger \cite{smirnov2017doppelganger} also relies on classification layers for doppelganger mining from the predicted classification scores. However, these methods  require classification parameters to be learned for class mining, which is intractable for large-scale classes and complex non-Euclidean matchers (\textit{e.g.} QAConv). In \cite{wang2017train}, all training classes are divided into subspaces by clustering on averaged class representations, and then mini batches are sampled within each subspace. This method requires a full forward pass of all the training data, and the clustering operation cannot easily be scaled up to large-scale classes. In \cite{harwood2017smart}, SmartMining was proposed, which builds an approximate nearest neighbor graph for all training samples after a full forward pass of the training data for feature extraction. However, this instance-level mining can be very expensive in computation, and even infeasible for complex non-Euclidean metric layers. In contrast, we propose and prove that sampling one example per class for class mining works well for large-scale deep metric learning without classification or instance-level mining.

  \section{Deep Metric Learning}
  There are two popular ways for learning deep person re-identification neural networks. The first one is the classification based method \cite{Zheng-CVPR2017-IDE}, also known as using the identification loss, or ID loss. This is a straightforward extension from general image classification. Since person re-identification is an open-class problem, the learned classifier is usually dropped after training. The last feature embedding layer is usually adopted instead (known as the identity embedding, or IDE \cite{Zheng-CVPR2017-IDE}), and the Euclidean or cosine distance is applied to measure the distance between two person images. The second one is the triplet loss based method \cite{Schroff2015-FaceNet,Hermans2017-Triplet}, which is usually combined with the ID loss. Together with the online hard example mining, the triplet loss is a very useful auxiliary loss function for enhancing the discriminability of the learned model.

  However, the above methods always require classifier parameters, which incur large memory and computational costs in both the forward and backward passes of large-scale deep learning. When dot products are employed for classification this is still acceptable to some extent. However, with more complex modules, e.g. QAConv \cite{Liao-ECCV2020-QAConv} where a full feature map convolution is required for matching, learning with class signatures is difficult to scale up.

  Therefore, for large-scale deep metric learning, we consider removing classification layers. Accordingly, pairwise matching between mini-batch samples is another solution \cite{yi2014deep,li2014cuhk03}. We adopt QAConv as our baseline method, which is the recent state of the art for generalizable person re-identification. It constructs query adaptive convolutional kernels on the fly for image matching, which suits pairwise learning. However, the original design of QAConv learning is based on the so-called class memory, which stores one feature map for each class for image-to-class matching, instead of using pairwise matching between mini-batch samples. Considering the matching complexity of the QAConv layer, this is not efficient in large-scale learning. Therefore, we only consider pairwise matching between mini-batch samples for QAConv, and remove its class memory.% module.% For convenience, the resulting method is denoted by QAConv-P.

  \section{Graph Sampling}

  \subsection{Motivation}
  As discussed, for deep metric learning, the well-known PK sampler \cite{Hermans2017-Triplet} is typically used to provide mini-batch samples. However, its random nature makes the sampled instances not informative enough for discriminant learning. In the PK sampler, as shown in Fig. \ref{fig:GS} (a), $P$ classes and $K$ images per class are randomly sampled for each mini batch. Though an online hard example mining (OHEM) was further proposed in \cite{Hermans2017-Triplet} to find informative instances within a mini batch% for learning
  , the PK sampler itself is still not efficient, as it provides limited hard examples for OHEM to mine.

  Therefore, the sampling method itself needs to be improved so as to provide informative samples for mini batches. Instead of using fully random sampling, the relationships among classes need to be considered. Thus, we construct a graph for all classes at the beginning of each epoch, and always sample nearest neighboring classes in a mini batch so as to enable discriminant learning. We call this idea graph sampling (GS), which is %introduced in detail in the following subsection.
  detailed below.

  \subsection{GS Sampler}

  At the beginning of each epoch, we utilize the latest learned model to evaluate the distances or similarities between classes, and then construct a graph for all classes. This way, the relationships between classes can be used for informative sampling. Specifically, we randomly select one image per class to construct a small sub-dataset. Then, the feature embeddings with the current network are extracted, denoted as $\mathbf{X}\in R^{C\times d}$, where $C$ is the total number of classes for training, and $d$ is the feature dimension. Next, %the current distance function (if parameterized) is used to calculate the 
  pairwise distances between all the selected samples are computed, e.g. by QAConv. As a result, a distance matrix $dist\in R^{C\times C}$ for all classes is obtained.

  Then, for each class $c$, the top $P-1$ nearest neighboring classes can be retrieved, denoted by $\mathcal{N}(c)=\{x_i|i=1,2,\ldots,P-1\}$, where $P$ is the number of classes to sample in each mini batch. Accordingly, a graph $G=(V,E)$ can be constructed, where $V=\{c|c=1,2,\ldots,C\}$ represents the vertices, with each class being one node, and $E=\{(c_1, c_2)|c_2 \in \mathcal{N}(c_1)\}$ represents the edges.

  Finally, for the mini-batch sampling, for each class $c$ as anchor, we retrieve all its connected classes in $G$. Then, together with the anchor class $c$, we obtain a set $A=\{c\}\bigcup\{x|(c,x)\in E\}$, where $|A|=P$. Next, for each class in $A$, we randomly sample $K$ instances per class to generate a mini batch of $B=P\times K$ samples for training. 
  A pseudocode of the GS sampler is shown in Appendix A.

  Note that, different from other mini-batch sampling methods, for the GS sampler the number of mini batches or iterations per epoch is always $C$, which is independent to the parameters $B$, $P$, and $K$. Nevertheless, the parameter $B$ still affects the computational load of each mini batch. Besides, one may worry that the GS sampler will be computationally expensive. However, note that, firstly, only one image per class is randomly sampled for the graph construction; and, secondly, the above computation is performed only once per epoch. In practice, we find that the GS sampler with QAConv, which is already a heavy matcher compared to the mainstream Euclidean distance, only requires tens of seconds for thousands of identities. Details will be presented in the experimental section.% \ref{sec:ablation-baseline}.

  \subsection{Loss Function}
  With mini batches provided by the GS sampler, we apply QAConv to compute similarity values between each pair of images, and formulate a triplet-based ranking learning problem within mini batches. Accordingly, we compute the batch OHEM triplet loss \cite{Hermans2017-Triplet} alone for metric learning:
  \begin{equation}\label{eq:loss}
   \begin{split}
  \ell(\boldsymbol{\theta}; X) = \sum_{i=1}^P \sum_{a=1}^K [m -& \min_{p=1\ldots K} s(f_{\boldsymbol{\theta}}(x_i^a), f_{\boldsymbol{\theta}}(x_i^p))\\
  +& \max_{\substack{j=1\ldots P \\ j \ne i \\ n=1\ldots K}} s(f_{\boldsymbol{\theta}}(x_i^a), f_{\boldsymbol{\theta}}(x_j^n)) ]_+,
   \end{split}
  \end{equation}
  where $X=\{x_i^a, i\in[1, P], a\in[1, K]\}$ contains the mini-batch samples, $\boldsymbol{\theta}$ is the network parameter, $f_{\boldsymbol{\theta}}$ is the feature extractor, $s(\cdot, \cdot)$ is the similarity, and $m$ is the margin. 
  
  Note that Eq. (\ref{eq:loss}) is usually used as an auxiliary to the ID loss, but not alone in person re-identification. This is probably because random samplers including PK cannot provide informative mini batches for OHEM to mine, which makes Eq. (\ref{eq:loss}) very small or even zero, and so the learning is not efficient. In contrast, with the proposed GS sampler, we prove that the OHEM triplet loss works well by itself.

  \subsection{Gradient Clipping}

  Note that the GS sampler already provides almost the hardest mini batches, and the  batch OHEM triplet loss further finds the hardest triplets within a mini batch for training. As a result, the model may suffer optimization difficulty, which in turn may impact convergence during training. In practice, we find that limiting $K=2$ alleviates this problem significantly. Or otherwise, the binary cross-entropy loss for pairwise matching can be a more stable alternative to the OHEM triplet loss (see Appendix B).

  Furthermore, to stabilize the training with the GS sampler and the hard triplet loss, we clip the gradient norm during the backward propagation. Specifically, let $\mathbf{g}$ be the gradient of all parameters, and $\|\mathbf{g}\|$ be its norm. The gradient will be clipped as $\mathbf{g} \leftarrow \min(1, \frac{T}{\|\mathbf{g}\|}) \cdot \mathbf{g}$, where $T$ is a predefined threshold. That is, if the gradient norm is larger than $T$ then clip it to be $T$. 
  Note that GS and OHEM provide the hardest examples, which facilitates discriminant learning. However, this may also lead to overfitting. Therefore, besides stabilizing the training, the gradient clipping operation is also useful to regularize noisy gradients to avoid overfitting on source
  domain, and, in turn, improving the generalization performance. The effect of this gradient clipping will be analyzed in the experiments.

  \section{Experiments}
  \subsection{Implementation Details}
  Our implementation is adapted from the official PyTorch code of QAConv \cite{Liao-ECCV2020-QAConv} (MIT license). We first build an improved baseline based on QAConv. Specifically, ResNet-50 \cite{he2016resnet} is used as the backbone, with IBN-b layers appended, following several recent studies \cite{pan2018two,jia2019frustratingly,Zhou2019-OSNet,Jin2020-SNR,Zhuang2020-CBN}. The layer3 feature map is used, with a neck convolution of 128 channels appended as the final feature map. The input image size is $384\times128$ (see Appendix E for results with $256\times128$). Several commonly used data augmentation methods are applied, including random cropping, flipping, occlusion \cite{Liao-ECCV2020-QAConv}, and color jittering. The batch size is set to 64. The SGD optimizer is adopted to train the network, with a learning rate of 0.0005 for the backbone, and 0.005 for newly added layers. The maximal learning epochs are 60. When the initial loss is reduced as a factor of 0.7, the learning rates are decayed by 0.1, and an early stopping is triggered after a further half of the already learned epochs. %This early stopping strategy usually results in less epochs required for learning, for example, less than 20 epochs when trained on MSMT17. 
  Gradient clipping is applied with $T=8$. Automatic Mixed Precision (AMP) in PyTorch is applied to accelerate training. When the proposed GS sampler is further applied (denoted by QAConv-GS), we use the hard triplet loss ($m$=16), instead of the class memory based loss proposed in \cite{Liao-ECCV2020-QAConv}, and the default parameters for GS are B=64, and K=2.

  \subsection{Datasets}
  Four large-scale person re-identification datasets, CUHK03 \cite{Li-CVPR-2014-DeepReID}, Market-1501 \cite{zheng2015smarket}, MSMT17~\cite{Wei-CVPR18-PTGAN}, and RandPerson \cite{Wang2020-RandPerson} are used in our experiments. The CUHK03 dataset contains 1,360 persons and 13,164 images. The most challenging subset named detected is used for our experiments. Besides, the CUHK03-NP protocol \cite{zhong2017re} is adopted, with 767 and 700 subjects used for training and testing, respectively. The Market-1501 dataset includes 32,668 images of 1,501 identities captured from six cameras. The training subset contains 12,936 images from 751 identities, while the test subset includes 19,732 images from 750 identities%, which is further divided into 3,368 query images and 15,913 gallery images
  . The MSMT17 dataset contains 4,101 identities and 126,441 images captured from 15 cameras. It is divided into a training set of 32,621 images from 1,041 identities, and a test set with the remaining images from 3,010 identities. %The test set further contains 11,659 query images and 82,161 gallery images.
  The RandPerson dataset is a recently released synthetic person re-identification dataset. It contains 8,000 persons and 1,801,816 images. We use the subset including 132,145 images of the 8,000 identities. This dataset is only used for large-scale training and generalization testing.
  Cross-dataset evaluation \cite{yi2014deep,Hu2014Cross} is performed on all datasets, by training on the training subset of one dataset (except that with MSMT17 we further used an additional setting with all images for training), and evaluating on the test subset of another dataset. Rank-1 and mean average precision (mAP) are used as the performance metrics, evaluated under single-query evaluation protocol.

  \subsection{Comparison to the State of the Art}

  A comparison to the state of the art (SOTA) in generalizable person re-identification is shown in Table \ref{tab:sota}, where three datasets are used for training, and three others are used for testing. Note that, with MSMT17 as the training set, one setting is to use all images for training, regardless of its subset splits. This is denoted by MSMT17 (all). Several generalizable person re-identification methods published recently are compared, including OSNet-IBN \cite{Zhou2019-OSNet}, OSNet-AIN \cite{zhou2021-osnet-ain}, MuDeep \cite{Qian2020-MuDeep}, SNR \cite{Jin2020-SNR}, QAConv \cite{Liao-ECCV2020-QAConv}, CBN \cite{Zhuang2020-CBN}, ADIN \cite{Yuan2020-ADIN}, and M$^3$L \cite{zhao2021learning}. Table \ref{tab:sota} shows that QAConv-GS significantly improves the previous SOTA. For example, with Market-1501 $\rightarrow$ CUHK03, the Rank-1 and mAP are improved by 8.8\% and 9.0\%, respectively. With Market-1501 $\rightarrow$ MSMT17, they are improved by 20.6\% and 7.7\%, respectively. With MSMT17 (all) $\rightarrow$ Market-1501, the improvements are 9.8\% for Rank-1 and 13.8\% for mAP. With RandPerson as the training data, the improvements on Market-1501 are 12\% for Rank-1 and 7.4\% for mAP, while the improvements on MSMT17 are 25.1\% for Rank-1 and 8.7\% for mAP. Though RandPerson is synthetic, the results show that models learned on it generalize quite well to real-world datasets, which confirms the findings in \cite{Wang2020-RandPerson}.

  \begin{table*}
  \centering
  \begin{tabular}{|c|c|c|c|c|c|c|c|c|}
    \hline
    \multirow{2}{*}{\tabincell{c}{Method}} & \multirow{2}{*}{\tabincell{c}{Venue}} & \multirow{2}{*}{\tabincell{c}{Training}} & \multicolumn{2}{c|}{CUHK03-NP} & \multicolumn{2}{c|}{Market-1501} & \multicolumn{2}{c|}{MSMT17} \\
    \cline{4-9}
     &  &  & Rank-1 & mAP & Rank-1 & mAP & Rank-1 & mAP \\
    \hline
    \hline
    M$^3$L \cite{zhao2021learning} & CVPR'21 & Multi & 33.1 & 32.1 & 75.9 & 50.2 & 36.9 & 14.7 \\
    \hline
    \hline
    MGN \cite{wang2018learning,Qian2020-MuDeep} &	ACMMM'18&	Market-1501&		8.5&	7.4& \cellcolor{lightgray} 95.7 & \cellcolor{lightgray} 86.9 &	-&	-\\
    MuDeep \cite{Qian2020-MuDeep} & TPAMI'20 & Market-1501  & 10.3 & 9.1 & \cellcolor{lightgray} 95.3 & \cellcolor{lightgray} 84.7 & - & - \\		
    QAConv \cite{Liao-ECCV2020-QAConv} & ECCV'20 & Market-1501  & 9.9 & 8.6 & \cellcolor{lightgray} - & \cellcolor{lightgray} - & 22.6	&7.0 \\
    OSNet-AIN \cite{zhou2021-osnet-ain}	&TPAMI'21	& Market-1501	&-	&-		& \cellcolor{lightgray} 94.2	& \cellcolor{lightgray} 84.4	&23.5	&8.2\\
    CBN \cite{Zhuang2020-CBN} & ECCV'20 & Market-1501  & - & - & \cellcolor{lightgray}  91.3 & \cellcolor{lightgray}  77.3 & 25.3 & 9.5 \\
  QAConv-GS&	Ours&	Market-1501 	&\textbf{19.1}	&\textbf{18.1}	& \cellcolor{lightgray} 91.6	& \cellcolor{lightgray} 75.5	&\textbf{45.9}	&\textbf{17.2}\\  % 25.07
  \hline
  \hline
  PCB \cite{sun2017pcb,Yuan2020-ADIN}	&ECCV'18&	MSMT17 &		-&	-&	52.7&	26.7& \cellcolor{lightgray} 	-& \cellcolor{lightgray} 	-\\
  MGN \cite{wang2018learning,Yuan2020-ADIN} &	ACMMM'18&	MSMT17&	-&	-&	48.7&	25.1& \cellcolor{lightgray} 	-& \cellcolor{lightgray} 	-\\
  ADIN \cite{Yuan2020-ADIN} &	WACV'20&	MSMT17&		-&	-&	59.1&	30.3& \cellcolor{lightgray} 	-& \cellcolor{lightgray} 	-\\
  SNR \cite{Jin2020-SNR}&	CVPR'20&		MSMT17&		-&	-&	70.1&	41.4& \cellcolor{lightgray} 	-& \cellcolor{lightgray} 	-\\
  CBN \cite{Zhuang2020-CBN} &	ECCV'20&	MSMT17&		-&	-&	73.7&	45.0& \cellcolor{lightgray} 72.8 & \cellcolor{lightgray} 42.9 \\
  QAConv-GS	&Ours	&MSMT17		&\textbf{20.9}	&\textbf{20.6}	&\textbf{79.1}	&\textbf{49.5}	& \cellcolor{lightgray} 79.2	& \cellcolor{lightgray} 50.9\\  % 42.51
  \hline
  \hline					
  OSNet-IBN \cite{Zhou2019-OSNet}	&CVPR'19	&MSMT17 (all)		&-	&-	&66.5	&37.2	& \cellcolor{gray} -	& \cellcolor{gray} -\\				
  OSNet-AIN \cite{zhou2021-osnet-ain}	&TPAMI'21	&MSMT17 (all)		&-	&-	&70.1	&43.3	& \cellcolor{gray} -	& \cellcolor{gray} -\\
  QAConv \cite{Liao-ECCV2020-QAConv}	&ECCV'20	&MSMT17 (all)		&25.3	&22.6	&72.6	&43.1	& \cellcolor{gray} -	& \cellcolor{gray} -\\
  QAConv-GS	&Ours	&MSMT17 (all)		&\textbf{27.6}	&\textbf{28.0}	&\textbf{82.4}	&\textbf{56.9}	& \cellcolor{gray} -	& \cellcolor{gray} -\\ % 48.71
  \hline
  \hline
  RP Baseline \cite{Wang2020-RandPerson} & ACMMM'20	&RandPerson	&13.4	&10.8	&55.6	&28.8	&20.1	&6.3\\
  CBN \cite{zhang2021unrealperson} &	ECCV'20 &	RandPerson &	- &	-&	64.7 &	39.3 &	20.0 &	6.8 \\
  QAConv-GS	&Ours	&RandPerson		&\textbf{18.4}	&\textbf{16.1}	&\textbf{76.7}	&\textbf{46.7}	&\textbf{45.1}	&\textbf{15.5}\\  % 36.41
  \hline
  \end{tabular}
  \caption{Comparison of the state-of-the-art direct cross-dataset evaluation results (\%). MSMT17 (all) means all images are used for training, regardless of subset splits. M$^3$L is trained on three datasets selected from CUHK03, Market-1501, DukeMTMC-reID, and MSMT17, while the other is held for testing. Results in gray cells are with within-dataset evaluation for a reference. ``-'' means not reported or not applicable.} \label{tab:sota}
  \end{table*}

  Note that, M$^3$L \cite{zhao2021learning} uses a different evaluation protocol, and thus the results are not directly comparable. Specifically, M$^3$L is trained on three datasets selected from CUHK03, Market-1501, DukeMTMC-reID\footnote{It is no longer available, so we do not use it in our experiments.}, and MSMT17, while the other is held for testing. Impressive results are obtained by M$^3$L on CUHK03-NP, which, though not directly comparable, exceed all our results, including those trained with all MSMT17 images. However, on Market-1501, the proposed method trained on MSMT17 outperforms M$^3$L in Rank-1 by 3.2\%, while the mAPs are comparable. Furthermore, on MSMT17, the proposed method trained on Market-1501 significantly outperforms M$^3$L, with 9\% gain in Rank-1 and 2.5\% in mAP. This is quite encouraging, since in both cases our training dataset is a subset of that used by M$^3$L.

  \subsection{Ablation Study}

  \subsubsection{Comparison of QAConv variants}\label{sec:ablation-baseline}

  Table \ref{tab:qaconv} shows a comparison among different variations of QAConv: the original QAConv \cite{Liao-ECCV2020-QAConv} (denoted as Ori), the competitive QAConv baseline we adapted (denoted as Base), and the proposed QAConv-GS. It shows that, beyond the successful learning scheme of the class memory module proposed in QAConv, QAConv-GS with the proposed GS sampler is also very effective in learning discriminant models. QAConv-GS outperforms the competitive baseline for all experiments, by 6.8\% and 5.4\% in Rank-1, respectively, on CUHK03 and Market-1501 when trained on MSMT17.
  
  \begin{table}
    \centering
    {\setlength{\tabcolsep}{0.8mm}%
    \begin{tabular}{|c|c|c|c|c|c|c|c|c|}
      \hline
        & \multicolumn{2}{c|}{Training} & \multicolumn{2}{c|}{CUHK03} & \multicolumn{2}{c|}{Market} & \multicolumn{2}{c|}{MSMT17} \\
      \cline{2-9}
       &   Data & Hours & R1 & mAP & R1 & mAP & R1 & mAP \\
      \hline
      \hline
      Ori  & Market & 1.33 & 9.9 & 8.6 & \cellcolor{lightgray}  - & \cellcolor{lightgray} - & 22.6	&7.0 \\
      Base&		Market	& 0.47	&14.6	&14.6& \cellcolor{lightgray} 88.7 & \cellcolor{lightgray} 71.4 &	42.6&	15.8\\
    GS&		Market &	\textbf{0.25}	&\textbf{19.1}	&\textbf{18.1}	& \cellcolor{lightgray} 91.6	& \cellcolor{lightgray} 75.5	&\textbf{45.9}	&\textbf{17.2}\\  % 25.07
    \hline
    \hline
    Base&		MSMT&	1.33	&14.1	&15.7	&73.7	&44.7	& \cellcolor{lightgray} 72.5	& \cellcolor{lightgray} 43.4 \\
    GS		&MSMT	& \textbf{0.73}	&\textbf{20.9}	&\textbf{20.6}	&\textbf{79.1}	&\textbf{49.5}	& \cellcolor{lightgray} 79.2	& \cellcolor{lightgray} 50.9\\  % 42.51
    \hline
    \hline	
    Ori		&MS-all	& 26.9	&25.3	&22.6	&72.6	&43.1	& \cellcolor{gray} -	& \cellcolor{gray} -\\
    Base		&MS-all	& 15.0	&23.4	&23.1	&80.1	&53.2	& \cellcolor{gray} -	& \cellcolor{gray} -\\
    GS	&MS-all	& \textbf{3.42}	&\textbf{27.6}	&\textbf{28.0}	&\textbf{82.4}	&\textbf{56.9}	& \cellcolor{gray} -	& \cellcolor{gray} -\\ % 48.71
    \hline
    \hline
    Base		&RP		& 25.4	&15.2	&14.6	&75.9	&46.0	&44.4	&\textbf{15.5}\\
    GS		&RP	& \textbf{2.0}	&\textbf{18.4}	&\textbf{16.1}	&\textbf{76.7}	&\textbf{46.7}	&\textbf{45.1}	&\textbf{15.5}\\  % 36.41
    \hline
    \end{tabular}}
    \caption{Comparison of QAConv variants. Ori: the original QAConv \cite{Liao-ECCV2020-QAConv}. Base: the competitive baseline we adapted. GS: graph sampling (ours). MS-all: MSMT17 (all). RP: RandPerson.}\label{tab:qaconv}
    \end{table}

  Interestingly, Table \ref{tab:qaconv} also shows that the within-dataset evaluation results are also improved by QAConv-GS compared to the baseline. However, improving performance on a single dataset does not always lead to better generalization, since it may also overfit a dataset, as can be observed in Table \ref{tab:sota}. Therefore, we suggest a focus on generalization since it is more critical for practical applications.

  Furthermore, we also compare the training time of QAConv (with class memory) and QAConv-GS. Both methods are tested on a single NVIDIA V100 GPU. From the comparison shown in Table \ref{tab:qaconv}, it can be observed that the original QAConv learned with class memory becomes very slow when trained on large-scale datasets, such as the full MSMT17 or RandPerson. This is not surprising, because in each mini-batch iteration, the QAConv with class memory needs to compute matching scores between mini-batch samples and the feature map memory of all classes; and the number of classes is 4,101 in MSMT17, and 8,000 in RandPerson. In contrast, the proposed pairwise learning with the GS sampler is much more efficient because it avoids matching all classes in each iteration. As can be seen from Table \ref{tab:qaconv}, the training time of the baseline QAConv can be reduced from 25.4 hours to 2 hours when trained on RandPerson with 8,000 identities, which is a significant achievement.

  In addition, we also evaluate the sampling efficiency of the proposed GS sampler. As stated earlier, it constructs a graph at the beginning of each epoch. We evaluate the running time of all the computations in GS. The results are 4 seconds on Market-1501, 9 seconds on the MSMT17 training subset, 40 seconds on the full MSMT17 dataset, and 138 seconds on RandPerson with 8,000 identities. Therefore, the GS sampler is in fact efficient, despite being incorporated into QAConv, which is a heavy matcher compared to the mainstream Euclidean distance.

  \subsubsection{Comparison of different sampling methods}

  In Table \ref{tab:sampling}, using the same QAConv and hard triplet loss, we compare three mini-batch sampling methods, including PK, a clustering based method \cite{wang2017train} (denoted as Cluster), and GS. Besides, the implementation of PK and Cluster follows GS, where the number of batches per epoch is determined by the number of classes. For \cite{wang2017train}, since k-means does not support non-Euclidean metric, we replace it with spectral clustering. The subspace parameter $M$ in \cite{wang2017train} is set to 10, after an optimization in [5, 50]. From Table \ref{tab:sampling}, we can see that PK performs the worst, due to its fully random nature, which does not provide enough hard examples in mini batches. Besides, we can see that, with the subspace clustering method proposed in \cite{wang2017train}, the performance is generally improved, thanks to the more informative mini batches sampled within each cluster. However, feature extraction from the whole training set and clustering of all classes are time consuming. In contrast, the proposed GS sampler is more efficient, since it only considers one example per class for the graph construction. Furthermore, GS also achieves the best performance, with improvements over Cluster of up to 4.7\% in Rank-1, and 3.4\% in mAP. We believe that clustering is less effective than graph based GS due to two reasons. First, only cluster centers may be surrounded by their dense neighbors, while others, especially boundary points (classes), may not be always with their full set of neighbors in the same cluster. Second, mini-batch classes need to be randomly sampled within a cluster, of which the operation may further miss out some nearest neighbors of each class. 
  
  Furthermore, in Appendix C, applications to two other baselines, OSNet \cite{Zhou2019-OSNet} and TransMatcher \cite{Liao-NeurIPS2021-TransMatcher}, also verify the generality of GS's advantage over PK. Besides, application to unsupervised domain adaptation (UDA) with GS in SpCL \cite{Ge2020-SpCL} is discussed in Appendix D, and a variant of GS using class centers is analyzed in Appendix E.

   \begin{table}
    \centering
    {\setlength{\tabcolsep}{0.8mm}%
    \begin{tabular}{|c|c|c|c|c|c|c|c|c|}
      \hline
      \multirow{2}{*}{\tabincell{c}{Method}} & \multicolumn{2}{c|}{Training} & \multicolumn{2}{c|}{CUHK03} & \multicolumn{2}{c|}{Market} & \multicolumn{2}{c|}{MSMT17} \\
      \cline{2-9}
      &   Data & Time & R1 & mAP & R1 & mAP & R1 & mAP \\
      \hline
      \hline
      PK&		Market	& \textbf{99} &17.9	&17.0& \cellcolor{lightgray} 	-&	 \cellcolor{lightgray} -&	43.3&	15.6\\ % 23.43
      Cluster &		Market	& 117 & 18.4	&17.3& \cellcolor{lightgray} 	-& \cellcolor{lightgray} 	-&	44.0&	15.8\\ % 23.87
      GS&	Market 	& 100 &\textbf{19.1}	&\textbf{18.1}	& \cellcolor{lightgray} -	& \cellcolor{lightgray} -	&\textbf{45.9}	&\textbf{17.2}\\  % 25.07
    \hline
    \hline
    PK&		MSMT		& \textbf{141} &18.6	& 18.8 &	75.7&	46.1& \cellcolor{lightgray} 	-& \cellcolor{lightgray} 	-\\  % 39.79
    Cluster&		MSMT	& 196 & 18.4	&19.2&	77.2&	47.6& \cellcolor{lightgray}  -	& \cellcolor{lightgray} 	-\\ % 40.62
    GS	&MSMT		& 145 &\textbf{20.9}	&\textbf{20.6}	&\textbf{79.1}	&\textbf{49.5}	& \cellcolor{lightgray} -	& \cellcolor{lightgray} -\\  % 42.51
    \hline
    \hline
    PK&		MS-all		& \textbf{669} &24.5	&24.6&	78.7&	52.1&	 \cellcolor{gray} -&	 \cellcolor{gray} -\\ % 44.98
    Cluster&		MS-all	& 881 & 26.3	&26.3&	80.4&	54.2&  \cellcolor{gray} -	&	  \cellcolor{gray} -\\ % 46.78
    GS		& MS-all	& 685 &\textbf{27.6}	&\textbf{28.0}	&\textbf{82.4}	&\textbf{56.9}	& \cellcolor{gray} -	& \cellcolor{gray} -\\ % 48.71
    \hline
    \hline
    PK &		RP		& \textbf{1,150} & 16.9	& 14.7 &	73.2 &	43.5 &	40.3 &	13.1 \\  % 33.63
    Cluster &		RP	& 1,922 & 17.3	& 15.0 &	73.3 &	43.3 &	40.4 &	13.4\\ % 33.79
    GS		& RP		& 1,397 &\textbf{18.4}	&\textbf{16.1}	&\textbf{76.7}	&\textbf{46.7}	&\textbf{45.1}	&\textbf{15.5}\\  % 36.41
    \hline
    \end{tabular}}
    \caption{Comparison of different sampling methods. MS-all: MSMT17 (all). RP: RandPerson. Time is with seconds per epoch.} \label{tab:sampling}
    \end{table}

  \subsubsection{Parameter analysis}

  \begin{figure}
  \centering
  \includegraphics[width=60mm]{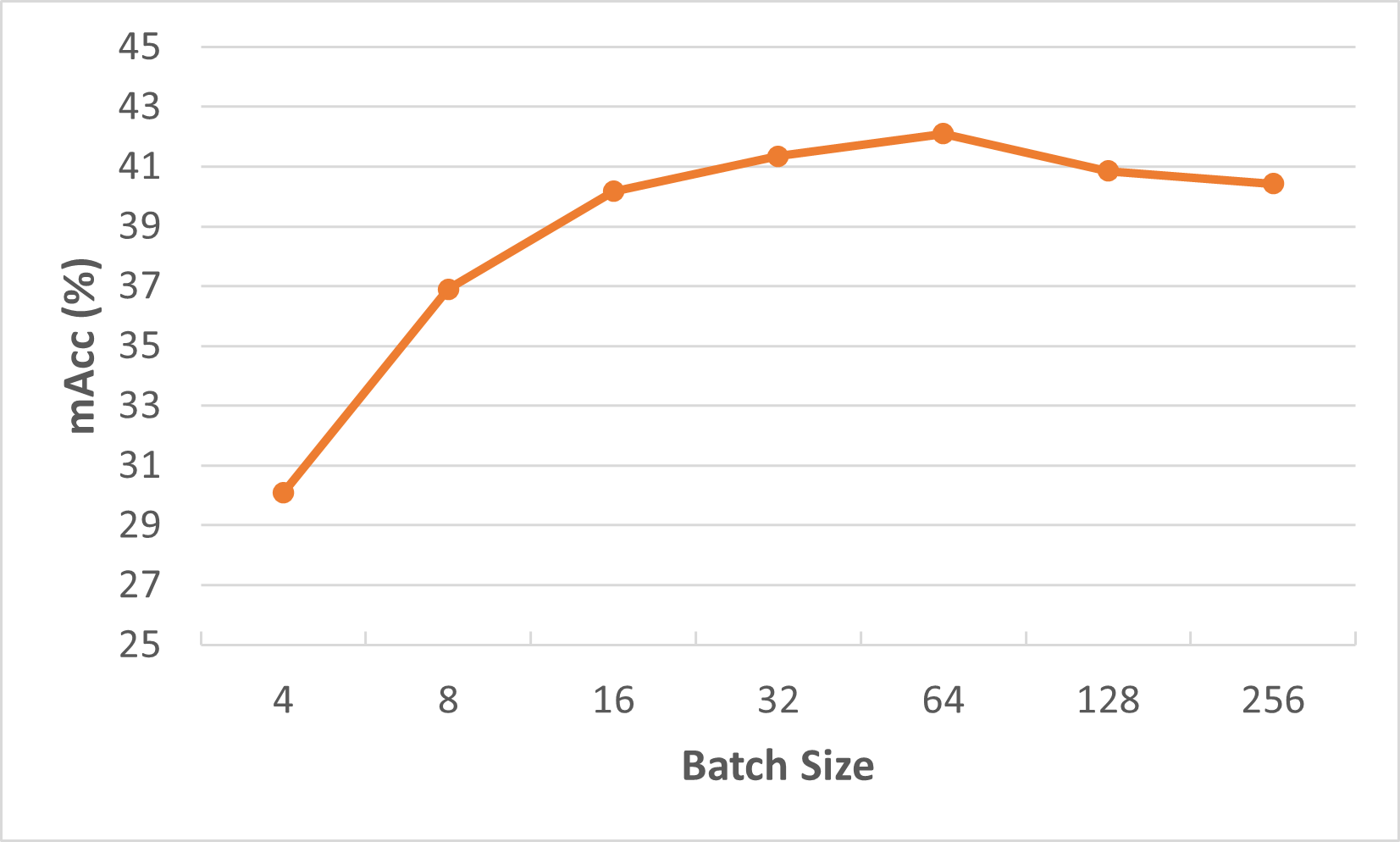}\\
  (a) Effect of batch size $B$\\
  \vspace{1mm}
  \includegraphics[width=60mm]{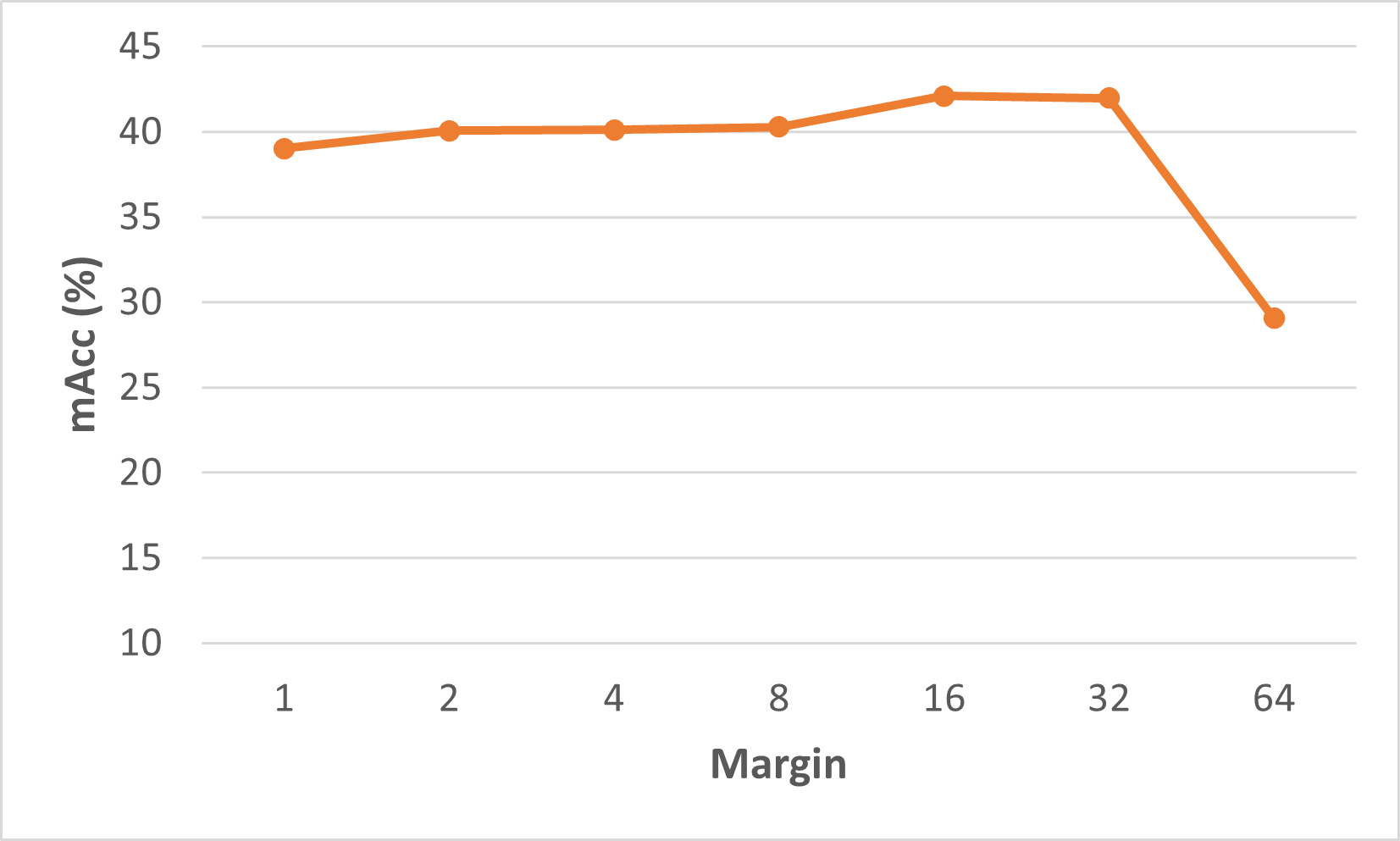}\\
  (b) Effect of margin $m$\\
  \caption{mAcc (\%) performance with (a) different batch sizes, and (b) different margin parameters, trained on MSMT17.}\label{fig:para-pk}
  \end{figure}

  In Fig. \ref{fig:para-pk}, we show the performance of the proposed method with different batch sizes and margin parameters. The training is performed on MSMT17. For ease and reliable comparison, we report the average (denoted by mAcc) of all Rank-1 and mAP results on all test sets over four random runs. We observe that, generally, the accuracy increases with increasing batch size $B$, but saturates at 64. As for the margin parameter $m$, note that the QAConv similarity score $s(\cdot, \cdot)$ used in Eq. (\ref{eq:loss}) ranges in (-$\infty$, +$\infty$). Fig. \ref{fig:para-pk}(b) shows that the performance slightly improves with increasing $m$ due to the increased discriminability, and achieves the best with $m$=16. However, after $m$=32, the performance drops significantly, due to intractable learning difficulty.

  \begin{figure}
  \centering
  \includegraphics[width=80mm]{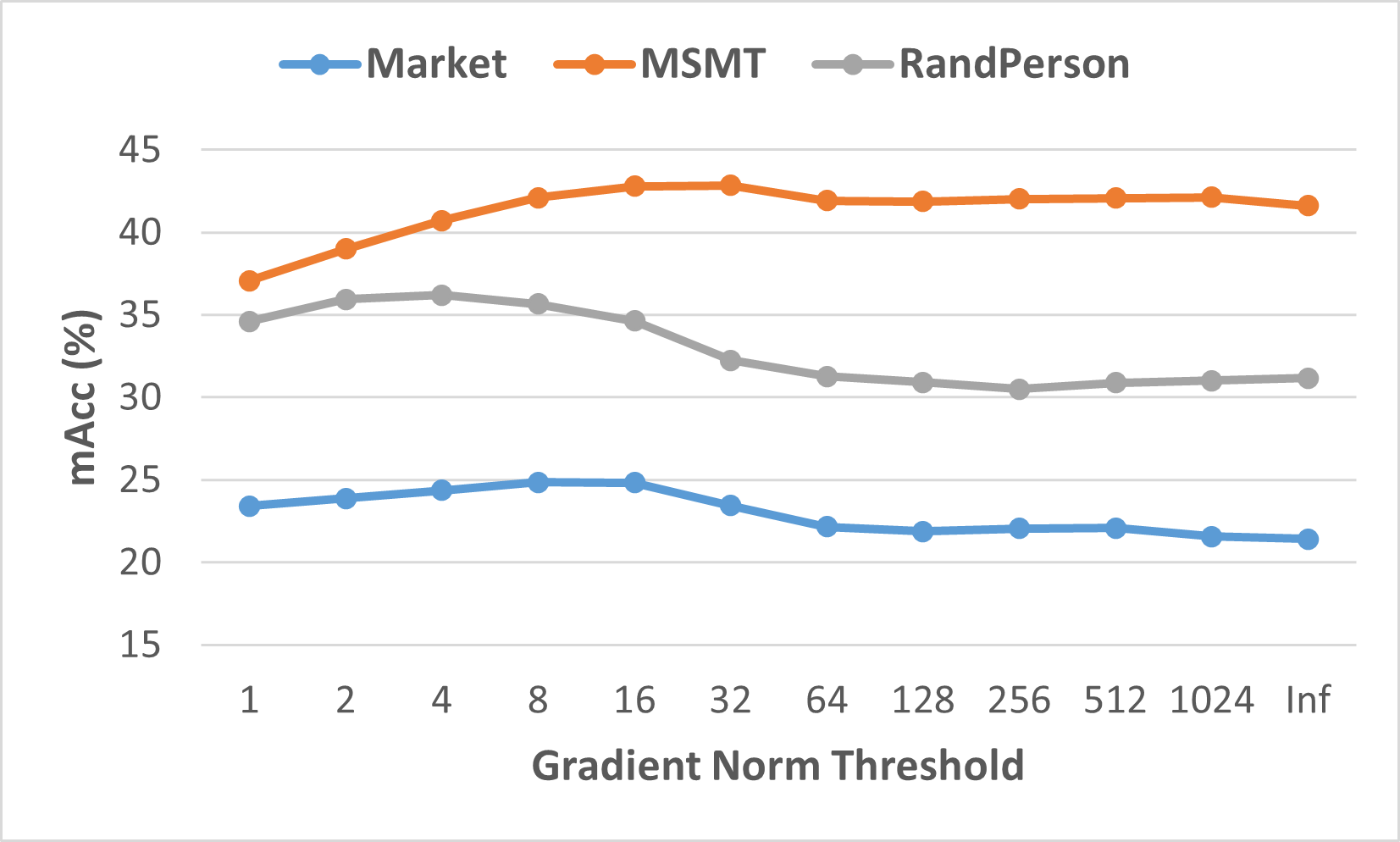}\\
  \caption{Influence of gradient clipping, trained on three datasets.}\label{fig:init-clip}
  \end{figure}

  \subsubsection{Effect of gradient clipping} \label{sec:ablation-clip}

  Next, we study the effect of gradient clipping on the learning of QAConv-GS. The results are shown in Fig. \ref{fig:init-clip}. Interestingly, when trained on MSMT17, the performance is less affected without gradient clipping (Inf). Specifically, with gradient clipping, only a slight improvement can be obtained, but too small threshold $T$ even prevent effective model learning. This is because, in our experiments, MSMT17 is the most comprehensive dataset. It provides large-scale and diverse training examples, which prevents overfitting in the view of ``regularization from data''. However, with the small-scale training dataset Market-1501, and the quite different synthetic dataset RandPerson, gradient clipping does provide useful regularization for model training, and improves the generalization performance. Therefore, a reasonable value of $T$=8 is considered as a trade-off.

  \begin{figure}[htb]
  \centering
  \includegraphics[width=30mm]{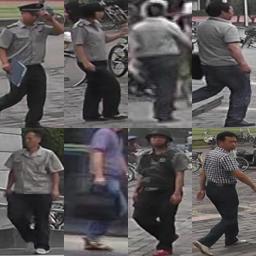}
  \hspace{0.1mm}
  \includegraphics[width=30mm]{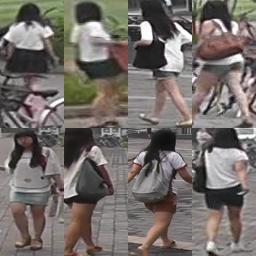}\\
  \vspace{1mm}
  \includegraphics[width=30mm]{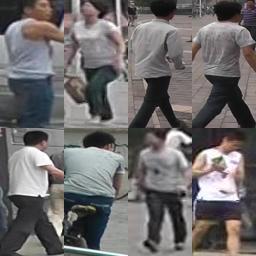}
  \hspace{0.1mm}
  \includegraphics[width=30mm]{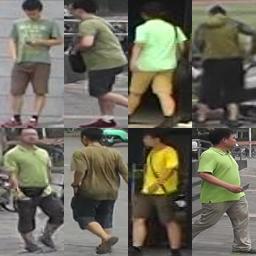}\\
  (a) Market-1501\\
  \vspace{2mm}
  \includegraphics[width=30mm]{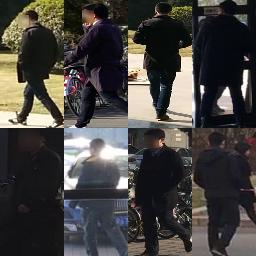}
  \hspace{0.1mm}
  \includegraphics[width=30mm]{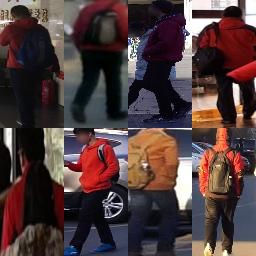}\\
  \vspace{1mm}
  \includegraphics[width=30mm]{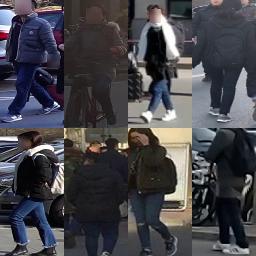}
  \hspace{0.1mm}
  \includegraphics[width=30mm]{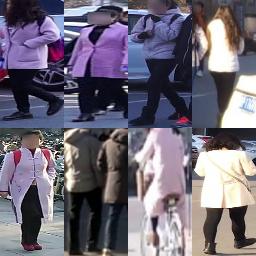}\\
  (b) MSMT17\\
  \caption{Groups of examples of the nearest neighboring classes generated by the GS sampler when trained on (a) Market-1501 and (b) MSMT17. In each group, the upper left image is the center class, and other images are the top-7 nearest neighboring classes.}\label{fig:gs_show}
  \end{figure}

  \subsubsection{Visualization of GS}

  Fig. \ref{fig:gs_show} shows some examples (more in Appendix F) of the nearest neighboring classes generated by the GS sampler. As can be observed, the GS sampler is indeed able to find similar classes as hard examples to challenge the learning. For example, it identifies similar kinds of clothes, colors, patterns, and accessories. These confusing examples help a lot in learning discriminative models.

  \section{Conclusion}
  With this study, we show that the popular PK sampler is not efficient in deep metric learning, and thus we propose a new batch sampler, called the graph sampler, to help learning discriminant models more efficiently. This is achieved by constructing a nearest neighbor graph of all classes for informative sampling. Together with a competitive baseline, we achieve the new state of the art in generalizable person re-identification with a significant improvement. Meanwhile, the training time is much reduced by removing the classification parameters and only using the pairwise distances between mini batches for loss computation. We believe the proposed technique is general and may also be applied in other fields, such as image retrieval, and face recognition, among others. Moreover, we discuss social impacts and some limitations of this research in Appendix G and H.

  \section*{Acknowledgements}
  Special thanks to Yanan Wang who helped creating Fig. \ref{fig:GS}, and Anna Hennig who helped proofreading the paper.

%%%%%%%%% REFERENCES
{\small
\bibliographystyle{ieee_fullname}
\bibliography{../../../Bib/Liao}
}

% \title{Appendix}

% \author{}

% \maketitle

\twocolumn[{ \centering \Large \bf Appendix \par \vspace{1cm} }]

\appendix

\renewcommand{\thefigure}{\Alph{figure}}
\renewcommand{\thetable}{\Alph{table}}
\renewcommand{\theequation}{\Alph{equation}}

%%%%%%%%% BODY TEXT
\section{Pseudocode of the GS sampler}

A pseudocode of the GS sampler is shown in Algorithm \ref{alg:GS}.

\begin{algorithm}
    %\SetAlgoRefName{} % no count number
    \caption{Graph Sampler}
    \label{alg:GS}
    \textbf{Input:} Data source $\mathbf{D}$, feature extractor $\mathbf{f}$, pairwise distance function $\mathbf{d}$, batch size $B$, number of instances per class $K$.\\
    \textbf{Output:} Sample iterator of the dataset $\mathbf{D}$.\\
    \textbf{Initialization:} $pids$: list of all class IDs; $index\_dict$: dictionary of list containing all sample indices of each class.\\
    \textbf{Procedure:}\\
    ~~index = []\\
    ~~for p in $pids$:\\
    ~~~~~~index.append(random.choice($index\_dict$[p], size=1)) ~~\textcolor[rgb]{0.00,0.50,0.25}{\# randomly select one sample per class}\\
    ~~dataset = $\mathbf{D}$(index) ~~\textcolor[rgb]{0.00,0.50,0.25}{\# construct a small sub-dataset}\\
    ~~X = $\mathbf{f}$(dataset) ~~\textcolor[rgb]{0.00,0.50,0.25}{\# extract features}\\
    ~~dist = $\mathbf{d}$(X, X) ~~\textcolor[rgb]{0.00,0.50,0.25}{\# calculate pairwise distance}\\
    ~~dist[i,i] = Inf ~~\textcolor[rgb]{0.00,0.50,0.25}{\# ignore the diagonal elements}\\
    ~~P = B / K ~~\textcolor[rgb]{0.00,0.50,0.25}{\# number of classes in a mini batch}\\
    ~~topk\_index = topk(-dist, size=P-1) ~~\textcolor[rgb]{0.00,0.50,0.25}{\# find nearest neighboring classes}\\
    ~~index = []\\
    ~~for p in shuffle($pids$):\\
    ~~~~~~index.extend(random.choice($index\_dict$[p], size=K)) ~~\textcolor[rgb]{0.00,0.50,0.25}{\# randomly select K samples per class}\\
    ~~~~~~for k in topk\_index[p]:\\
    ~~~~~~~~~~index.extend(random.choice($index\_dict$[k], size=K)) ~~\textcolor[rgb]{0.00,0.50,0.25}{\# randomly select K samples per class}\\
    \textbf{Return:} iter(index)
\end{algorithm}

\section{Alternative Loss Function and Analysis}\label{sec:binary-loss}
\subsection{Binary Cross Entropy Loss}

Note that the batch hard triplet loss (Eq. (1) in the main paper) is usually used as an auxiliary to the classification loss, but not alone, in person re-identification. This is probably because random samplers including PK cannot provide informative mini batches for OHEM to mine, which makes Eq. (1) very small or even zero, and so the learning is not efficient. In contrast, with the proposed GS sampler, we prove that the OHEM triplet loss works well by itself with $K=2$. We use this loss function alone because, as motivated in the main paper, we aim at removing classification layers for large-scale metric learning.

However, note that the GS sampler already provides almost the hardest mini batches, and the batch hard triplet loss further finds the hardest triplets within a mini batch for training. As a result, the model may suffer optimization difficulty, which in turn may impact convergence during training. In practice, we find that limiting $K=2$ alleviates this problem significantly, while $K>2$ usually makes the learning not able to converge.

Alternatively, pairwise verification or binary classification is another solution \cite{yi2014deep,li2014cuhk03} for pairwise matching or metric learning within mini batches. Specifically, we apply QAConv to compute similarity values between a pair of images, and formulate a pairwise verification or binary classification problem in mini-batch based learning. Accordingly, we compute the binary cross entropy loss as follows.
\begin{equation}\label{eq:bce-loss}
\ell(\boldsymbol{\theta}) = -\frac{1}{B}\sum_{i=1}^{B}\sum_{j\ne i} y_{ij} log(p_{ij}(\boldsymbol{\theta})) + (1 - y_{ij}) log(1 - p_{ij}(\boldsymbol{\theta})),\\
\end{equation}
where $B$ is the mini-batch size, $\boldsymbol{\theta}$ is the network parameter, $p_{ij}\in[0,1]$ is the QAConv similarity indicating binary classification probability, and $y_{ij}=1$ indicates a positive pair, while a negative pair otherwise. By default, we choose $B=64$ and $K=4$ for this loss.

\subsection{Experimental Comparison}

Table \ref{tab:loss} shows a comparison between the hard triplet loss and the binary cross entropy loss for QAConv-GS. Results shown in the table indicate that, the hard triplet loss performs better than the binary cross entropy loss for all datasets, thanks to OHEM which further mines hard examples within mini batches provided by GS. However, the hard triplet loss used alone in the proposed pipeline is sensitive to $K$ values as discussed. In contrast, the binary cross entropy loss is a more stable alternative, working well with different $B$ and $K$ values. This will be analyzed in the following subsection.

\begin{table}
   \centering
   \begin{tabular}{|@{~}c@{~}|@{~}c@{~}|c@{~}|@{~}c@{~}|c@{~}|@{~}c@{~}|c@{~}|@{~}c@{~}|}
     \hline
     \multirow{2}{*}{\tabincell{c}{Train}}  & \multirow{2}{*}{\tabincell{c}{Method}} & \multicolumn{2}{|c|}{CUHK03} & \multicolumn{2}{|c|}{Market} & \multicolumn{2}{|c|}{MSMT17} \\
     \cline{3-8}
      &    & R1 & mAP & R1 & mAP & R1 & mAP \\
     \hline
     \hline
     \multirow{2}{*}{\tabincell{c}{Market}} &	Binary&	{16.4}	&{15.7}	&-	&-	&{41.2}	&{15.0}\\
   	 &	Triplet&	\textbf{19.1}	&\textbf{18.1}	&-	&-	&\textbf{45.9}	&\textbf{17.2}\\  % 25.07
   \hline
   \hline
   \multirow{2}{*}{\tabincell{c}{MSMT}}	&Binary	& {20.0}	&{19.2}	&{75.1}	&{46.7}	&-	&-\\
     & Triplet		&	\textbf{20.9}	&\textbf{20.6}	&\textbf{79.1}	&\textbf{49.5}	&-	&-\\  % 42.51
   \hline
   \hline	
   \multirow{2}{*}{\tabincell{c}{MS-all}}		&Binary &{27.2}	&{27.1}	&{80.6}	&{55.6}	&-	&-\\
    	& Triplet	&	\textbf{27.6}	&\textbf{28.0}	&\textbf{82.4}	&\textbf{56.9}	&-	&-\\ % 48.71
   \hline
   \hline
   \multirow{2}{*}{\tabincell{c}{RP}}	&Binary & {14.8}	&{13.4}	&{74.0}	&{43.8}	&{42.4}	&{14.4}\\
     	&Triplet		&\textbf{18.4}	&\textbf{16.1}	&{76.7}	&{46.7}	&\textbf{45.1}	&{15.5}\\  % 36.41
   \hline
   \end{tabular}
   \caption{Comparison of loss functions. Binary: binary cross entropy loss. Triplet: hard triplet loss. Market: Market-1501 dataset. MSMT: MSMT17 dataset. MS-all: MSMT17 (all). RP: RandPerson dataset.}\label{tab:loss}
   \end{table}

   \subsection{Parameter analysis}

   \begin{figure}
   \centering
   \includegraphics[width=80mm]{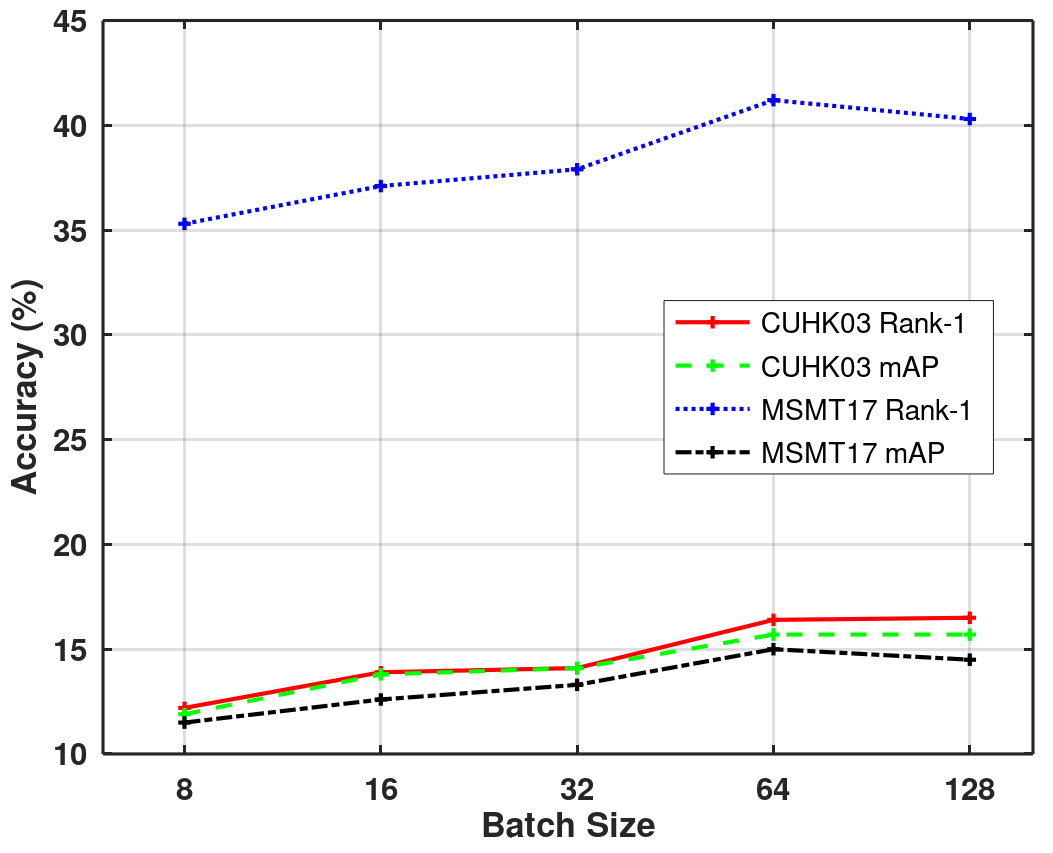}\\
   (a) Effect of batch size\\
   \vspace{3mm}
   \includegraphics[width=80mm]{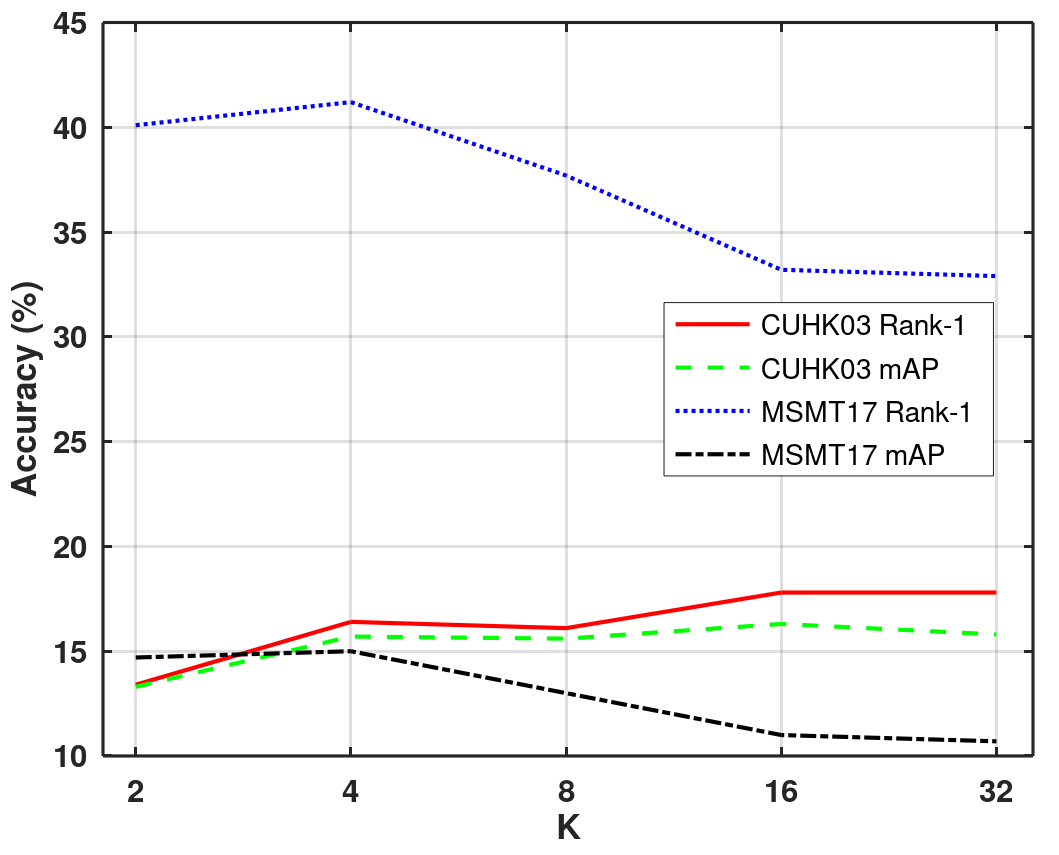}\\
   (b) Effect of $K$\\
   \caption{Performance with different parameter configurations of the GS sampler when the binary cross entropy loss is applied, trained on Market-1501. (a) with varying batch size under fixed $K=4$; and (b) with varying $K$ under fixed $B=64$.}\label{fig:para-pk}
   \end{figure}

   When the binary cross entropy loss is applied, in Fig. \ref{fig:para-pk}, we show the performance with different parameter configurations of the GS sampler, trained on Market-1501. We observe that for the batch size (Fig. \ref{fig:para-pk} (a)), generally the accuracy increases with the increasing batch size (thus increasing $P$), but saturates at about 64. It is understood that mini batches with larger batch size provides more comprehensive data for learning, however, at the cost of enlarged computation time, recalling that the number of iterations per epoch is fixed as $C$ for the GS sampler. For example, with $B=64$, the training of QAConv-GS on Market-1501 is about 0.68 hours on a single V100 GPU. However, this is about 1.32 hours with $B=128$ for training the same epochs.

   Next, we evaluate the influence of $K$ under fixed $B=64$, as shown in Fig. \ref{fig:para-pk} (b). Interestingly, larger $K$ leads to gradually better performance on the CUHK03-NP, however, it degrades the performance significantly on MSMT17. It appears that $K=4$ is a reasonable trade-off.

   Since the hard triplet loss performs better, in the following, by default we still use this loss.

\section{Application to Other Baselines}

Furthermore, to show the generality of the proposed graph sampling method, we apply it to two other algorithms, namely OSNet \cite{Zhou2019-OSNet} and TransMatcher \cite{Liao-NeurIPS2021-TransMatcher}.

The official code of OSNet\footnote{\url{https://github.com/KaiyangZhou/deep-person-reid}} (MIT License) is used. We used its osnet\_ibn\_x1\_0 config, with softmax+triplet loss and the PK sampler (RandomIdentitySampler) for the best performance, denoted by OSNet-IBN + PK. This combination of softmax+triplet loss and the PK sampler is also the most popular setting in person re-identification for strong baselines. Then, upon this baseline, we apply the proposed graph sampling to replace the PK sampler, denoted by OSNet-IBN + GS.
The training is performed on the MSMT17 (all), as in \cite{Zhou2019-OSNet}, and the learned models are evaluated on the CUHK03-NP and Market-1501 datasets. The results are shown in Table \ref{tab:osnet}. From the comparison it can be seen that the proposed GS sampler can also improve other strong baselines in replacing the popular PK sampler. Therefore, it is proved to be general and may also be applied to other methods.

\begin{table}[htb]
\centering
\caption{Direct cross-dataset evaluation results (\%) with different baselines trained on MSMT17 (all).}\label{tab:osnet}
\begin{tabular}{|@{}c@{}|c|c|c|c|}
  \hline
  \multirow{2}{*}{\tabincell{c}{Method}} & \multicolumn{2}{c|}{CUHK03} & \multicolumn{2}{c|}{Market}\\
  \cline{2-5}
   &  R1 & mAP & R1 & mAP \\
  \hline
  OSNet-IBN \cite{Zhou2019-OSNet} & - & - & 66.5 & 37.2 \\
  OSNet-IBN + PK & 23.4 & 23.6 & 67.9 & 39.6\\
  OSNet-IBN + GS	&\textbf{24.5}	&\textbf{24.9}	&\textbf{71.3}	&\textbf{42.6}\\
\hline
\end{tabular}
\end{table}

Furthermore, with a very recent method TransMatcher \cite{Liao-NeurIPS2021-TransMatcher}, we also compare the PK and GS samplers. The official code of TransMatcher \footnote{\url{https://github.com/ShengcaiLiao/QAConv/tree/master/projects/transmatcher}} (MIT License) is used, with its default settings. The results are shown in Table \ref{tab:TransMatcher}. It can also be observed that on average the proposed GS sampler performs much better than the PK sampler, verifying again the generality of GS.

\begin{table*}
  \centering
  \begin{tabular}{|c|c|c|c|c|c|c|c|c|}
    \hline
    \multirow{2}{*}{\tabincell{c}{Method}} & \multirow{2}{*}{\tabincell{c}{Training}} & \multicolumn{2}{c|}{CUHK03-NP} & \multicolumn{2}{c|}{Market-1501} & \multicolumn{2}{c|}{MSMT17} \\
    \cline{3-8}
     &  & Rank-1 & mAP & Rank-1 & mAP & Rank-1 & mAP \\
    \hline
    \hline
    TransMatcher-PK & Market-1501  & \textbf{22.9} & \textbf{21.5} & \cellcolor{lightgray} - & \cellcolor{lightgray}  - & 45.6 & 17.8 \\
    TransMatcher-GS &	Market-1501 	&22.2	&21.4	& \cellcolor{lightgray} -	& \cellcolor{lightgray} -	&\textbf{47.3}	&\textbf{18.4}\\
  \hline
  \hline
  TransMatcher-PK  &	MSMT17&	23.6 & \textbf{22.9} &	78.3 &	51.7 & \cellcolor{lightgray} - & \cellcolor{lightgray} - \\
  TransMatcher-GS	&MSMT17		&\textbf{23.7}	&22.5	&\textbf{80.1}	&\textbf{52.0}	& \cellcolor{lightgray} -	& \cellcolor{lightgray} -\\
  \hline
  \hline
  TransMatcher-PK 	&MSMT17 (all)		&30.7	&29.5	&79.9	&55.7	& \cellcolor{gray} -	& \cellcolor{gray} -\\
  TransMatcher-GS &MSMT17 (all)		&\textbf{31.9}	&\textbf{30.7}	&\textbf{82.6}	&\textbf{58.4}	& \cellcolor{gray} -	& \cellcolor{gray} -\\
  \hline
  \hline
  TransMatcher-PK  &	RandPerson &	\textbf{18.0} &	\textbf{16.5} &	73.3 &	45.3 &	40.6 &	14.1 \\
  TransMatcher-GS	&RandPerson		&17.1	&16.0	&\textbf{77.3}	&\textbf{49.1}	&\textbf{48.3}	&\textbf{17.7}\\
  \hline
  \end{tabular}
  \caption{Comparison of direct cross-dataset evaluation results (\%) using TransMatcher \cite{Liao-NeurIPS2021-TransMatcher} with PK and GS. MSMT17 (all) means all images are used for training, regardless of subset splits.} \label{tab:TransMatcher}
  \end{table*}

  \section{Application to Unsupervised Domain Adaptation}

  As a new scenario, we tried unsupervised domain adaptation (UDA) by replacing PK with GS in the source domain of SpCL \cite{Ge2020-SpCL}. The Rank1/mAP results for PK and GS are 86.1/70.9 and 87.3/71.5, respectively, for CUHK03-NP$\rightarrow$Market-1501. Slight improvements can be observed. However, for the time being, it is still not yet straightforward to apply GS for pseudo labeled samples by clustering on the target domain. This may be because pseudo labels could be noisy, and GS may be  aggressive in finding hard negative samples that could possibly be positive. To address this, further developments may be required in considering how to handle noisy samples, which is quite interesting.

\section{Further Ablation Studies}

\begin{table*}
  \centering
  \begin{tabular}{|c|c|c|c|c|c|c|c|c|}
    \hline
    \multirow{2}{*}{\tabincell{c}{Method}} & \multirow{2}{*}{\tabincell{c}{Training}} & \multicolumn{2}{c|}{CUHK03-NP} & \multicolumn{2}{c|}{Market-1501} & \multicolumn{2}{c|}{MSMT17} \\
    \cline{3-8}
     &  & Rank-1 & mAP & Rank-1 & mAP & Rank-1 & mAP \\
    \hline
    \hline
  QAConv-GS&	Market-1501 	&\textbf{19.1}	&\textbf{18.1}	& \cellcolor{lightgray} -	& \cellcolor{lightgray} -	&\textbf{45.9}	&\textbf{17.2}\\  % 25.07
  QAConv-GS ($256\times128$) & Market-1501  & 16.9 & 17.2 & \cellcolor{lightgray} - & \cellcolor{lightgray}  - & 45.4 & 17.1 \\
  QAConv-GS-Center & Market-1501  & 15.4 & 14.7 & \cellcolor{lightgray} - & \cellcolor{lightgray}  - & 45.0 & 15.7 \\
  \hline
  \hline
  QAConv-GS	&MSMT17		&\textbf{20.9}	&\textbf{20.6}	&\textbf{79.1}	&{49.5}	& \cellcolor{lightgray} -	& \cellcolor{lightgray} -\\  % 42.51
  QAConv-GS ($256\times128$)  &	MSMT17&	18.6 & 19.8 &	77.9 & \textbf{49.6}	& \cellcolor{lightgray} - & \cellcolor{lightgray} - \\
  QAConv-GS-Center  &	MSMT17&	15.3 &	16.1 &	73.9 &	41.5 & \cellcolor{lightgray} - & \cellcolor{lightgray} - \\
  \hline
  \hline
  QAConv-GS	&MSMT17 (all)		&\textbf{27.6}	&\textbf{28.0}	&\textbf{82.4}	&\textbf{56.9}	& \cellcolor{gray} -	& \cellcolor{gray} -\\ % 48.71
  QAConv-GS ($256\times128$) 	&MSMT17 (all)		& 24.3	&	25.6 & 81.5 	& 55.3	& \cellcolor{gray} -	& \cellcolor{gray} -\\
  QAConv-GS-Center 	&MSMT17 (all)		&	25.2 &	24.6 &	78.6 &	51.2 & \cellcolor{gray} -	& \cellcolor{gray} -\\
  \hline
  \hline
  QAConv-GS	&RandPerson		&\textbf{18.4}	&\textbf{16.1}	&{76.7}	&{46.7}	&\textbf{45.1}	&{15.5}\\  % 36.41
  QAConv-GS ($256\times128$)  &	RandPerson &	16.2 & 14.4 &	74.7 &	45.5  &	45.0  &	 \textbf{15.8}  \\
  QAConv-GS-Center  &	RandPerson &	17.4 &	15.4 &	\textbf{76.8}  &	\textbf{47.0}  &	44.3  &	15.2  \\
  \hline
  \end{tabular}
  \caption{Comparison of the direct cross-dataset evaluation results (\%) for different variants of QAConv-GS. QAConv-GS-Center is based on selecting class centers for graph construction for GS. MSMT17 (all) means all images are used for training, regardless of subset splits.} \label{tab:ablation}
  \end{table*}

In this paper, all experiments are with images of $384\times128$ as inputs. To understand the influence of image size, we also conduct experiments with images of $256\times128$ as inputs. The results are shown in Table \ref{tab:ablation}. It can be seen that the results are quite close to each other on Market-1501 and MSMT17, though results with $384\times128$ are clearly better than that of $256\times128$ on CUHK03. Note that our results with $256\times128$ still achieve the state of the art compared to existing methods in Table 1 of the main paper.

In the proposed GS, one example per class is sampled for the graph construction, which is efficient. Alternatively, class centers can also be considered for graph construction. In fact, class centers are used in \cite{wang2017train} for clustering based batch sampling, and we show better performance of GS in Table 3 of the main paper. To further understand this, we use class centers to construct graphs for GS. This is denoted by QAConv-GS-Center. The comparison results are shown in Table \ref{tab:ablation}. It is clear that GS performs better.

There might be two problems with class centers. First, it lacks flexibility of sample relationships, since many classes may have large distribution variances. This is also discussed in \cite{suh2019stochastic}. Second, computing class centers requires feature extraction of all training samples, which hinders large-scale learning. The average training time increases from 1.68 hours of QAConv-GS to 2.27 hours of QAConv-GS-Center. Especially, QAConv-GS costed 3.4 hours to train MSMT17 (all), but QAConv-GS-Center costed 5.4 hours.

\section{Visualization of GS}

\begin{figure*}
\centering
\includegraphics[width=27mm]{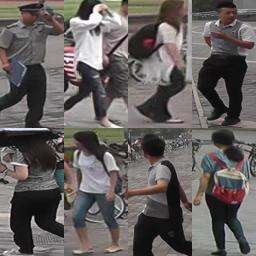}
\hspace{0.3mm}
\includegraphics[width=27mm]{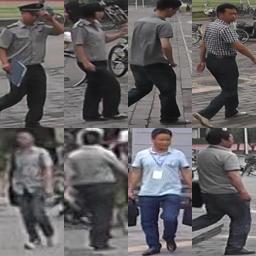}
\hspace{0.3mm}
\includegraphics[width=27mm]{gs_show/market/0077_ep15_c4s1_010951_01.jpg}
\hspace{3mm}
\includegraphics[width=27mm]{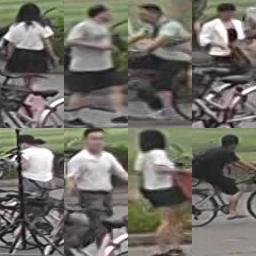}
\hspace{0.3mm}
\includegraphics[width=27mm]{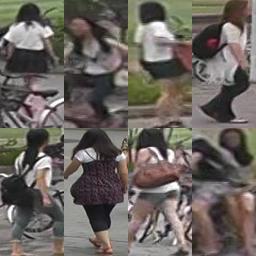}
\hspace{0.3mm}
\includegraphics[width=27mm]{gs_show/market/0107_ep15_c5s1_020001_02.jpg}\\
Market, epoch 2 \hspace{4mm} Market, epoch 8 \hspace{4mm} Market, epoch 15 \hspace{7mm} Market, epoch 2 \hspace{4mm} Market, epoch 8 \hspace{4mm} Market, epoch 15\\
\vspace{1mm}
\includegraphics[width=27mm]{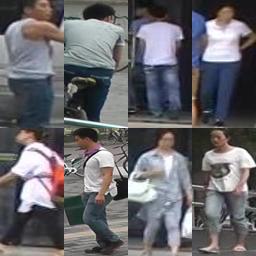}
\hspace{0.3mm}
\includegraphics[width=27mm]{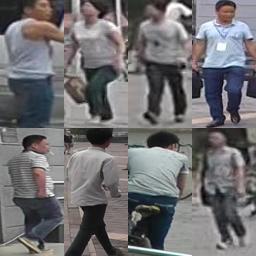}
\hspace{0.3mm}
\includegraphics[width=27mm]{gs_show/market/0067_ep15_c2s1_007851_01.jpg}
\hspace{3mm}
\includegraphics[width=27mm]{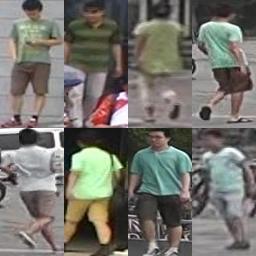}
\hspace{0.3mm}
\includegraphics[width=27mm]{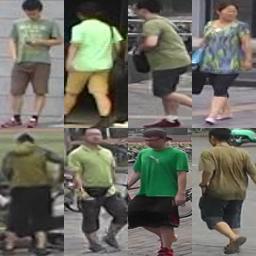}
\hspace{0.3mm}
\includegraphics[width=27mm]{gs_show/market/0068_ep15_c6s1_011901_01.jpg}\\
Market, epoch 2 \hspace{4mm} Market, epoch 8 \hspace{4mm} Market, epoch 15 \hspace{7mm} Market, epoch 2 \hspace{4mm} Market, epoch 8 \hspace{4mm} Market, epoch 15\\
\vspace{1mm}
\includegraphics[width=27mm]{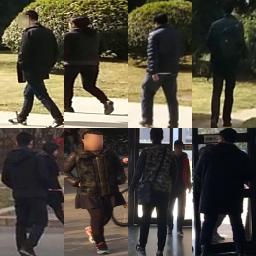}
\hspace{0.3mm}
\includegraphics[width=27mm]{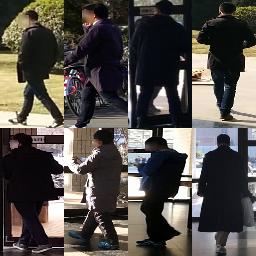}
\hspace{0.3mm}
\includegraphics[width=27mm]{gs_show/msmt/0113_ep15_023_10_0303noon_0275_0.jpg}
\hspace{3mm}
\includegraphics[width=27mm]{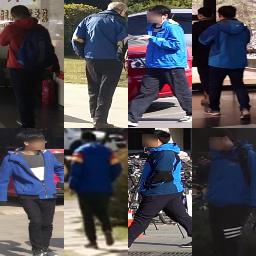}
\hspace{0.3mm}
\includegraphics[width=27mm]{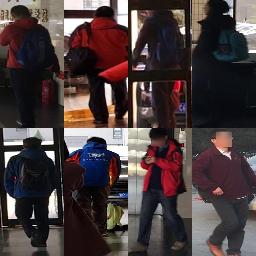}
\hspace{0.3mm}
\includegraphics[width=27mm]{gs_show/msmt/0127_ep15_001_13_0303noon_0621_1_ex.jpg}\\
MSMT, epoch 2 \hspace{4mm} MSMT, epoch 8 \hspace{4mm} MSMT, epoch 15 \hspace{6mm} MSMT, epoch 2 \hspace{4mm} MSMT, epoch 8 \hspace{4mm} MSMT, epoch 15\\
\vspace{1mm}
\includegraphics[width=27mm]{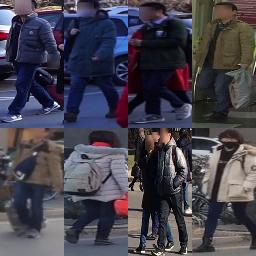}
\hspace{0.3mm}
\includegraphics[width=27mm]{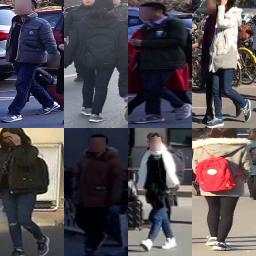}
\hspace{0.3mm}
\includegraphics[width=27mm]{gs_show/msmt/0031_ep15_016_14_0303morning_0643_1.jpg}
\hspace{3mm}
\includegraphics[width=27mm]{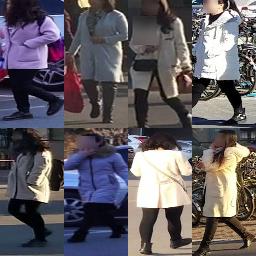}
\hspace{0.3mm}
\includegraphics[width=27mm]{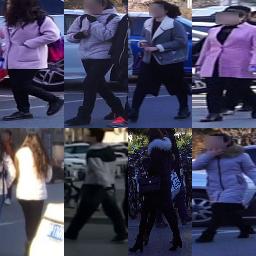}
\hspace{0.3mm}
\includegraphics[width=27mm]{gs_show/msmt/0057_ep15_022_14_0303morning_1142_2_ex.jpg}\\
MSMT, epoch 2 \hspace{4mm} MSMT, epoch 8 \hspace{4mm} MSMT, epoch 15 \hspace{6mm} MSMT, epoch 2 \hspace{4mm} MSMT, epoch 8 \hspace{4mm} MSMT, epoch 15\\
\caption{Eight groups of examples for the nearest neighboring classes generated by the GS sampler. The first two rows are from the training on Market-1501, while the last two rows are from that of MSMT17. In each group, three sets of images are shown, corresponding to epoch 2, 8, and 15. In each set, the upper left image is the center class, and other images are the top-7 nearest neighboring classes to the center class.}\label{fig:gs_show}
\end{figure*}

Finally, we show some examples for the nearest neighboring classes generated by the GS sampler in Fig. \ref{fig:gs_show}. It can be observed that, the GS sampler is indeed able to find similar classes as hard examples to challenge the learning. For example, similar kind of clothes, similar colors, patterns, and accessories. These confusing examples helps a lot in learning discriminative models. Besides, it seems that in early epochs, the model tends to evaluate similarity with visual appearance, regardless of the influence of foreground and background. However, in late epochs, the model learns to remove the influence of background, and learns higher level of abstraction. For example, in the upper right group, the similarity is less affected by bicycles in background with epoch 15. In the first group of MSMT17, the similarity is less affected by trees in background with later epochs. With epoch 15 of the first group, the model learns the concept of security guards. In the upper right group, with epoch 15 the model learns the concept of girls with short skirts. In the last group of Market-1501, with epoch 15 the clothes are more consistent in style and color. In the upper right group of MSMT17, with epoch 15 in GS the model correctly retrieves red coats. In the last group of MSMT17, with epoch 15 in GS the model correctly retrieves pink coats as well.

\section{Limitations}

The proposed method, despite achieving very good results, may have two limitations. First, GS requires additional computation for mini batch sampling. We design two ways to reduce the computation, that is, employing GS only at the beginning of each epoch, and randomly sampling only one sample per class for the distance computation and graph construction. As a result, the additional running time introduced by GS is still acceptable, as reported in Section 5.4.1 of the main paper. Besides, note that with GS the number of training epochs is generally reduced. For example, with GS the proposed method usually requires less than 20 epochs for training, while existing methods typically require 60 epochs or more to train. Therefore, GS deserves the additional computational costs. However, in our experiments, the maximal number of classes is only 8,000. GS may still have a big limitation with millions of identities, which need further investigation.

Second, as discussed, GS provides challenging examples for training, and so the default hard triplet loss only works well with $K=2$. Otherwise, the training is too difficult to converge. Nevertheless, as discussed in Section \ref{sec:binary-loss}, this limitation can be solved by employing the binary cross entropy loss as an alternative, though with inferior performance.

\section{Social Impacts}\label{sec:impacts}

Person re-identification is a technique to automatically search persons from a large amount of videos. It has potential social values in some practical applications, such as person image retrieval of suspects, character recognition in movies \cite{huang2020movienet}, and so on. For example, it is very useful to reduce large amount of human labors and greatly advance the effort in criminal investigation. Accordingly, person re-identification methods are actively studied. The person re-identification technique, however, may also be used by company for a surveillance of employees, or by malls for tracking of daily visitors. Therefore, it requires effective legislation to avoid abuse of this technique. This paper focuses on foundational research; it is not tied to particular applications, let alone deployments.

Besides, the research and developments of such technique are often with datasets collected from surveillance videos that may contain personally identifiable information. To address this, a positive action is to remove such information, as done in MSMT17v2 \cite{Wei-CVPR18-PTGAN} with facial areas masked. More promisingly, a better way recently demonstrated is to use synthesized data, as done in RandPerson \cite{Wang2020-RandPerson}.

%%%%%%%%% REFERENCES
{\small
\bibliographystyle{ieee_fullname}
\bibliography{../../../Bib/Liao}
}

\section*{Biography}

Shengcai Liao is the Director of Research and Developments (Acting) in the Inception Institute of Artificial Intelligence (IIAI), Abu Dhabi, UAE. He is a Senior Member of IEEE. Previously, he was an Associate Professor in the Institute of Automation, Chinese Academy of Sciences (CASIA). He received the B.S. degree in mathematics from the Sun Yat-sen University in 2005 and the Ph.D. degree from CASIA in 2010. He was a Postdoc in the Michigan State University during 2010-2012. His research interests include object detection, recognition, and tracking, especially face and person related tasks. He has published over 100 papers, with \textbf{15,800+ citations and h-index 44} according to Google Scholar. He \textbf{ranks 905 among 215,114 scientists (Top 0.42\%)} in 2019 single year in the field of AI, according to a study by Stanford University of Top 2\% world-wide scientists. His representative work LOMO+XQDA, known for effective feature design and metric learning for person re-identification, has been \textbf{cited 2,000+ times and ranks top 10 among 602 papers in CVPR 2015}. He was awarded/co-awarded the Best Student Paper in ICB 2006, ICB 2015, and CCBR 2016, and the Best Paper in ICB 2007. He was also awarded the IJCB 2014 Best Reviewer and CVPR 2019/2021 Outstanding Reviewer. He was an Assistant Editor for the book “Encyclopedia of Biometrics (2nd Ed.)”. He serves as Program Chair for IJCB 2022. He served as Area Chairs for ICPR 2016, ICB 2016 and 2018, and CVPR 2022, SPC for IJCAI 2021, and reviewers for ICCV, CVPR, ECCV, NeurIPS, ICLR, AAAI, TPAMI, IJCV, TIP, etc. His team was the Winner of the CVPR 2017 Detection in Crowded Scenes Challenge and ICCV 2019 NightOwls Pedestrian Detection Challenge. Homepage: \url{https://liaosc.wordpress.com/}

\end{document}